\pdfoutput=1
\documentclass{article}
\usepackage[nonatbib, final]{icbinb_neurips_2020}
\usepackage[numbers]{natbib}

\usepackage[utf8]{inputenc} 
\usepackage[T1]{fontenc}    
\usepackage{hyperref}       
\usepackage{url}            
\usepackage{booktabs}       
\usepackage{amsfonts}       
\usepackage{nicefrac}       
\usepackage{microtype}      
\usepackage{float}
\usepackage{dsfont}
\usepackage[binary-units]{siunitx}

\usepackage{amsmath,amsfonts,bm}

\def\eqref#1{equation~\ref{#1}}

\def\1{\bm{1}}

\DeclareMathAlphabet{\mathsfit}{\encodingdefault}{\sfdefault}{m}{sl}
\SetMathAlphabet{\mathsfit}{bold}{\encodingdefault}{\sfdefault}{bx}{n}

\newcommand*{\B}[1]{\ifmmode\bm{#1}\else\textbf{#1}\fi}
\newcommand*{\I}[1]{\ifmmode\mathit{#1}\else\textit{#1}\fi}
\renewcommand*{\t}[1]{\ifmmode\text{#1}\else #1 \fi}

\newcommand\m{\ensuremath{\B{m}}}
\newcommand\s{\ensuremath{\B{s}}}
\newcommand\z{\ensuremath{\B{z}}}
\newcommand\x{\ensuremath{\B{x}}}
\newcommand*\N{\ensuremath{\mathcal{N}}}

\def\note[#1#2#3]{#1\if b#2$\flat_#3$\else\if#2##$\sharp_#3$\else$_#2$\fi\fi}

\usepackage[export]{adjustbox}
\usepackage{xcolor}
\usepackage{graphicx}
\graphicspath{{graphics/}}
\usepackage{subcaption}
\usepackage[tableposition=top]{caption}
\usepackage{algorithm}
\usepackage{algpseudocode}

\usepackage{tikz}
\usetikzlibrary{shapes,arrows,fit,chains,matrix,backgrounds}
\usetikzlibrary{positioning, shapes.geometric}
\tikzstyle{node} = [circle, minimum size = 18mm, thick, draw =black!80]
\tikzstyle{nodeinter} = [rectangle, minimum size = 16mm, thick, draw =black!80, fill=gray!30]
\tikzstyle{nodeinterwhite} = [rectangle, minimum size = 16mm, thick, draw =black!80]
\tikzstyle{nodeobserved} = [circle, minimum size = 18mm, thick, draw =black!80, fill=gray!30]
\tikzstyle{box} = [rectangle, draw =black!0]
\tikzstyle{arrow} = [thick,->,>=stealth,line width=0.6mm]
\tikzstyle{arrow2} = [dashed,->,>=stealth]

\definecolor{purple}{rgb}	{0.69, 0.38, 0.53}
\definecolor{orange}{rgb}	{0.84, 0.36, 0.05}
\definecolor{blue}{rgb}	{0.27, 0.52, 0.53}
\definecolor{green}{rgb}	{0.6, 0.59, 0.1}
\definecolor{red}{rgb}	{0.8, 0.14, 0.11}
\definecolor{yellow}{rgb}	{0.84, 0.6, 0.13}
\definecolor{aqua}{rgb}	{0.41, 0.62, 0.42}
\definecolor{extlinkcolor}{rgb}	{0.03, 0.4, 0.47}
\definecolor{intlinkcolor}{rgb}	{0.69, 0.23, 0.01}

\title{Problems using deep generative models for probabilistic audio source separation}

\author{%
  Maurice Frank\\
  Amsterdam Machine Learning Lab\\
  University of Amsterdam\\
  \texttt{maurice.frank@posteo.de} \\
   \And
   Maximilian Ilse \\
   Amsterdam Machine Learning Lab\\
  University of Amsterdam\\
   \texttt{m.ilse@uva.nl} \\
}

\begin{document}

\maketitle

\begin{abstract}
Recent advancements in deep generative modeling make it possible to learn prior distributions from complex data that subsequently can be used for Bayesian inference. However, we find that distributions learned by deep generative models for audio signals do not exhibit the right properties that are necessary for tasks like audio source separation using a probabilistic approach. We observe that the learned prior distributions are either discriminative and extremely peaked or smooth and non-discriminative. We quantify this behavior for two types of deep generative models on two audio datasets.
\end{abstract}

\section{Introduction: Langevin dynamics for source separation}%
\label{sec:introduction}
Our initial goal was to use Langevin dynamics~\cite{neal_mcmc_2012} in combination with deep generative priors to perform source separation of audio mixes. Our approach closely follows the work of~\cite{jayaram_source_2020}, where mixed images are successfully separated. The biggest advantage of the approach used by~\cite{jayaram_source_2020} is that it does not rely on pairs of source signals and mixes as required by SOTA audio source separation models~\cite{narayanaswamy_audio_2019, kaspersen_hydranet_2019}. Last, in contrast to~\cite{narayanaswamy_unsupervised_2020}, we are interested in performing the source separation in the time domain, which has multiple advantages like decreased computational complexity as well as the preservation of the phase of the signals. As common in the source separation literature, we assume that the mix $\pmb{m} \in \mathcal{X}$ is a linear combination of $N$ source signals $\pmb{s}_1, \dots, \pmb{s}_N \in \mathcal{X}$
\begin{align}
    \pmb{m} = g(\pmb{s}) = \sum_{k=1}^N \alpha_k \pmb{s}_k.
\end{align}
As seen in~\cite{jayaram_source_2020}, we take a probabilistic approach in order to solve the source separation problem. According to Bayes rule we can compute the posterior distribution using
\begin{align}
    p(\pmb{s}|\pmb{m}) = \frac{p(\pmb{s})p(\pmb{m}|\pmb{s})}{p(\pmb{m})},
\end{align}
where we use a Gaussian approximation $p(\pmb{m}) = \mathcal{N}(g(\pmb{s}), \gamma^2 I)$ with noise parameter $\gamma$. Stochastic Gradient Langevin Dynamics (SGLD)~\cite{welling_bayesian_2011} enables us sample from the posterior distribution $p(\pmb{s}|\pmb{m})$ without the need for evaluating $p(\pmb{m})$. A new sample of the source signals can be generated by
\begin{align}
    \pmb{s}^{t+1} = \pmb{s}^t + \eta \nabla_{\pmb{s}} \left(\log p(\pmb{s}^t) + \frac{1}{2\gamma^2}\left\lVert \pmb{m} - g(\pmb{s}^t) \right\rVert^2\right) + \sqrt{2 \eta} \epsilon,
    \label{eq:sgld}
\end{align}
where $\epsilon=\mathcal{N}(0, 1)$. Last, we assume that the prior of the source signals factorizes as follows
\begin{align}
    p(\pmb{s}) = p(\pmb{s}_1, \dots, \pmb{s}_N) = p(\pmb{s}_1)\dots p(\pmb{s}_N).
\end{align}
We choose to parameterize the priors $p(\pmb{s}_1)\dots p(\pmb{s}_N)$ with deep generative models. This allows us to easily compute $\nabla_{\pmb{s}} \log p(\pmb{s}^t)$ using the automatic differentiation tools from~\cite{paszke_pytorch_2019}.
As noted in~\cite{jayaram_source_2020}, in order to successfully recover the original source signals using Equation \ref{eq:sgld}, we require the priors $p(\pmb{s}_1)\dots p(\pmb{s}_N)$ to be discriminative as well as sufficiently smooth. After our initial experiments failed, we observed that the learned prior distributions are either discriminative and extremely peaked or smooth and non-discriminative. In the following, we quantify this behavior for two types of deep generative models on two audio dataset.

\section{Method: Modeling raw audio with generative models}%
\label{sec:method}
Deep learning models as used for image applications are unsuitable for raw audio signals (signals in the time-domain). Digital audio is sampled at high sample rates, commonly 16kHz up to 44kHz. The features of interest lie at scales of strongly different magnitudes. Therefore, generative models need to model the complete range of frequencies containing high-frequency features like timbre and slow frequency features like song structure. We will be using two types of generative models to learn the likelihood $p(\pmb{s})$ of audio signals, namely, WaveNet~\cite{van_den_oord_wavenet_2016} and FloWaveNet
\cite{kim_flowavenet_2019}.

\paragraph{WaveNet}
The WaveNet is an autoregressive generative model for raw audio. The generation of a new sample $x_t$ is conditioned on all previous samples $x_1, \dots, x_{t-1}$. The likelihood of the entire signal is given by
\begin{align}
    p(\pmb{x}) = \prod_{t=1}^T p(x_t|x_1, \dots, x_{t-1}).
\end{align}
After training, starting from a single sample the model is able to iteratively generate a coherent time signal. The distribution \(p(x_t|x_1, \dots, x_{t-1})\) is modeled as a multinomial logistic regression, therefore the continuous signal $x$ is discretized using a so-called \(\mu\)-law encoding~\cite{van_den_oord_wavenet_2016}, resulting in 256 classes. Due to the autoregressive nature of the Wavenet the genrative process is diffiult to parallalize and generally slow.

The WaveNet adapts the PixelCNN~\cite{van_den_oord_conditional_2016} architecture to the audio domain. It is a fully-convolutional network where dilated causal convolutions~\cite{yu_multi-scale_2016} are used. Using a stack of dilated convolutions increases the receptive field of the deep features without increasing the computational complexity. Further, the convolutions are gated~\cite{hochreiter_long_1997} and the output is constructed as the sum of skip connections from each layer. The skip connections fuse information from multiple time-scales.

\paragraph{FloWaveNet}
An alternative to using an autoregressive model to model \(p(\pmb{x})\) are normalizing flows~\cite{rezende_variational_2016}. Normalizing flows are a class of exact likelihood models, which are amenable to gradient-based optimisation and efficient in inference and sampling. A standard normalizing flow in continuous space, is based on the simple change of variables formula. Given an observed data variable \(\pmb{x} \in X\); a prior probability distribution \(p_Z(\cdot)\) on a latent variable \(\pmb{z} \in Z\), and a differentiable, bijective function \(\pmb{z} = f(\pmb{x})\), we can model a probability distribution \(p_X({\pmb{x}})\) as
\begin{align}\label{nf}
    p_X(\pmb{x})&=p_Z(f(\pmb{x}))\bigg\lvert \det\bigg(\frac{\partial f(\pmb{x})}{\partial\pmb{x}} \bigg) \bigg\rvert.
\end{align}

Since Equation~\ref{nf}  requires the computation of a determinant, a special architecture is necessary to reduce the computational cost. One such architecture was introduced in~\cite{dinh_density_2017}.~\cite{dinh_density_2017} is using so-called coupling layers in which an input \(\pmb{x}\) is masked into two equally sized parts \(\pmb{x}_a\) and \(\pmb{x}_b\). One part is fed into a function to provide the weights of an affine transformation ($s(\cdot), t(\cdot)$) of the other

\begin{align}
    \hat{\pmb{x}}_b = s(\pmb{x}_a) \cdot \pmb{x}_b + t(\pmb{x}_a).
\end{align}
The resulting Jacobian is a triangular matrix, whose determinant can be easily computed. There exist multiple approaches to combine the dilated convolutions used in the WaveNet and normalizing flows. In the case of FloWaveNet ~\cite{kim_flowavenet_2019} a WaveNet encoder is used to predict the weights of the affine transformation within every coupling layer. This architecture enables fast, parallel generation of audio samples.

\section{Experiments}%
\subsection{Datasets}%
\label{sec:experiments}%
\begin{figure}[h]
    \begin{minipage}{0.47\textwidth}
            \resizebox{\textwidth}{!}{%
            \begin{tikzpicture}[ampersand replacement=\&]
                \node[matrix,thick,column sep=1em,row sep=1em]
                {
                    \draw (0,0.5) sin (0.25,1) cos (0.5,0.5) sin (0.75,0) cos (1,0.5); \&
                    \draw (0,0) -- (1,1) -- (1,0);\&
                    \draw (0,0) -- (0,1) -- (0.5,1) -- (0.5,0) -- (1,0);\&
                    \draw (0,0) -- (0.5,1) -- (1,0); \\
                };
            \end{tikzpicture}
            }%
        \caption{One period of each of the four toy waveforms: sine, sawtooth, square and triangle wave.}%
        \label{fig:toy_data}%
    \end{minipage}\hfill%
    \begin{minipage}{0.47\textwidth}%
        \resizebox{\textwidth}{!}{%
            \begin{tikzpicture}[ampersand replacement=\&]
    \node[matrix,thick,column sep=1em,row sep=1em]
    {
        \node (bass)    {\includegraphics[width=35pt]{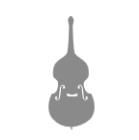}}; \&
        \node (drums){\includegraphics[width=35pt]{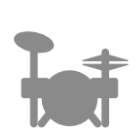}}; \&
        \node (voice)  {\includegraphics[width=35pt]{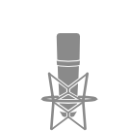}}; \&
        \node (other)  {\includegraphics[width=35pt]{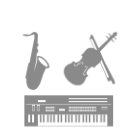}}; \\
    };
\end{tikzpicture}
        }%
        \caption{The four source channels for the \texttt{musdb18} dataset: bass ,drums, vocals and \I{other}.}%
        \label{fig:musdb_data}%
    \end{minipage}%
    \end{figure}

\paragraph{Toy data}
We create a toy-like dataset consisting of four distinct waveforms, as shown in Figure~\ref{fig:toy_data}: a sine, a sawtooth, a square and a triangle wave. We generate the waveforms using a sampling frequency of \(16 \si{\kHz}\). For each waveform we sample a random frequency $f\in [27 \si{\Hz}, 4186 \si{\Hz}]$, a random amplitude $A\in[0.8, 1.0]$ and a random phase $\phi \in[0, 2\pi]$.
\begin{align}\label{eq:mean}
    \pmb{m} = \frac{1}{4} \sum_i^4 \pmb{s}_i
\end{align}
The mix $\pmb{m}$ is equal to the mean of the four source signals. We create 5000 mixes of one second length for training and a testset of 1500 mixes.

\paragraph{musdb18}
The \texttt{musdb18}~\cite{rafii_musdb18_2017} dataset, created for the 2018 Signal Separation Evaluation Campaign~\cite{stoter_2018_2018}, is a benchmark dataset used to evaluate audio source separation algorithms. The dataset consists of 150 songs from various artists and genres, split into train and test sets sized 100 and 50, respectively. For each song, the full mix $\pmb{m}$ and four separate sources $\pmb{s}_1, \pmb{s}_2, \pmb{s}_3, \pmb{s}_4$ are given:  \emph{drums}, \emph{bass}, \emph{vocals} and \emph{others}. The \emph{others} source contains any instruments not contained in the first three. Note that the mix $\pmb{m}$ does not strictly follow Equation~\ref{eq:mean} since it involves audio effects like compression. The song files are encoded with a sampling rate of \(44.1 \si{\kHz}\) which we down-sample to \(16 \si{\kHz}\). We extract 150 fixed-length frames of one second from each song.

\subsection{Discrimative power of the priors}
\label{sec:discriminative}
\begin{figure}
    \centering
    \begin{subfigure}{0.22\textwidth}
    \includegraphics[width=\textwidth]{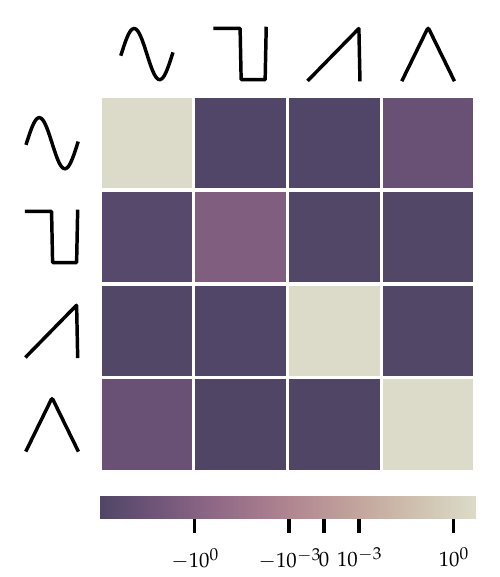}%
    \caption{FloWaveNet}%
    \label{fig:noiseless_channels_toy}%
\end{subfigure}
\begin{subfigure}{0.22\textwidth}
    \includegraphics[width=\textwidth]{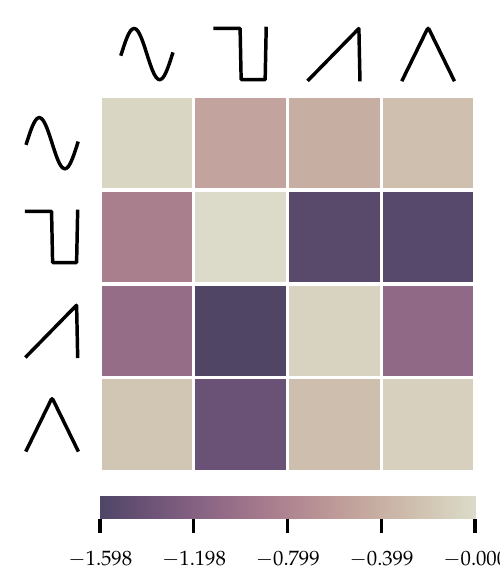}%
    \caption{WaveNet}%
    \label{fig:noiseless_channels_toy_wn}%
\end{subfigure}
\hfill
\begin{subfigure}{0.22\textwidth}
    \includegraphics[width=\textwidth]{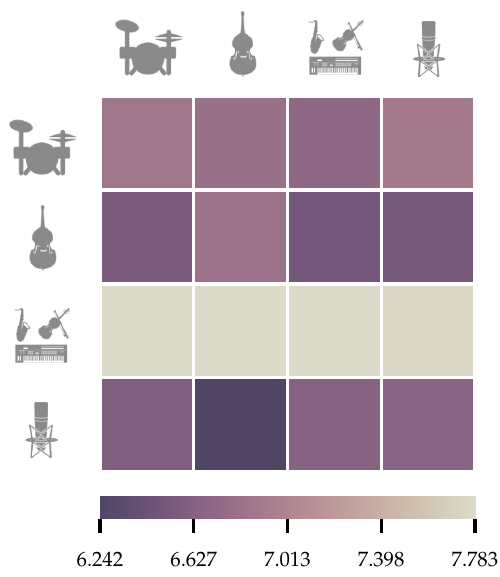}%
    \caption{FloWaveNet}%
    \label{fig:noiseless_channels_musdb}%
\end{subfigure}
\begin{subfigure}{0.22\textwidth}
    \includegraphics[width=\textwidth]{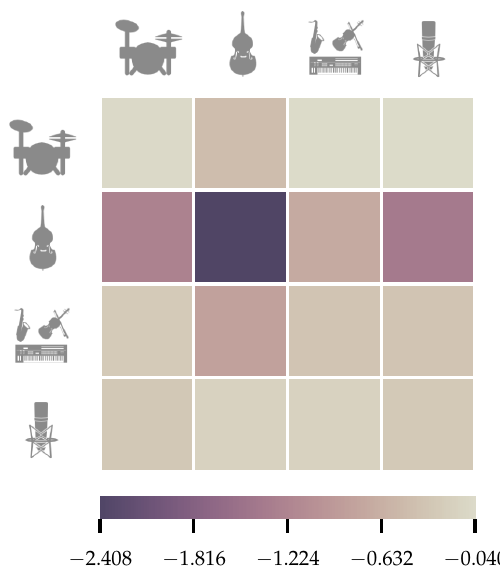}%
    \caption{WaveNet}%
    \label{fig:noiseless_channels_musdb_wn}%
\end{subfigure}
    \caption{The log-likelihood of each source under each prior for both sets of priors. Notice how for the Toy data we get the diagonal that we are expecting while for the real music the likelihood for every field is high and in the same range. The \I{other} prior assigns the highest likelihoods.}%
    \label{fig:noiseless_channels}%
\end{figure}

In the following, we train a separate WaveNet and FloWaveNet model for each signal source type of both datasets, in total eight separate generative models for each dataset. Details about  architecture choices and training schedules can be found in the Appendix~\ref{sec:appendix}.

Using Langevin dynamics for separation we optimize the separated source frames under each prior model. During training of the deep generative priors, they explicitly contract the density for positive, in-class examples. During separation, the priors encounter negative out-of-distribution samples for the first time. To be useful for separation, the priors have to give a low likelihood to samples from the other possible sources.

Following the results in~\cite{nalisnick_deep_2019}, we test the out-of-distribution detection performance of the deep generative priors by evaluating the mean log-likelihood of the test data of each source under each source prior. In Figure~\ref{fig:noiseless_channels} we show that only for the FloWaveNet model trained with the Toy data behaves as anticipated. The in-class samples have a high likelihood while for all prior models the out-of-class samples have a low likelihood.

For the \texttt{musdb18} dataset, neither the WaveNet- nor the FloWaveNet-based priors can discriminate between in-class and out-of-class samples. We hypothesize that this stems from the fact that the real musical data is severely more complicated compared than the Toy data. The in-class variability of the real sources is that high, that the models are not able to learn a distribution that would be discriminative. Note that the source \textit{other} in \texttt{musdb18} contains an undefined set of instruments making a model of those sounds in general impractical. But even if we are ignoring this subset neither prior model can discriminate the remaining source types. As a result, for the following experiments we focus on the Toy data dataset.

\subsection{Smoothness of the learned distribution}
In the case of the Toy dataset, only the FloWaveNet priors can distinguish between in-class and out-of-class signals. However, when we tried to use those priors for source separation as described in Section~\ref{sec:introduction}, we failed. We argue that one possible explanation is the peakedness of the learned prior distributions. The probability mass learned by the model is peaked at true samples but quickly decays with more disturbance of the input. The reason for this behavior is that all models are trained with noise-free samples of their respective signal sources (sine, square, saw, triangle).

As proposed in~\cite{jayaram_source_2020}, we now approximate the noisy distribution \(\log p_{\sigma}(\B{x})\), which is the convolution of the noiseless distribution with a Gaussian with variance \(\sigma\): \(\log p_{\sigma}(\B{x}) * \N(0, \sigma)\) by adding Gaussian noise with the same variance to the input. Figuratively speaking, the Gaussian noise in data space translates to Gaussian smoothing of the peak in the probability distribution of the data.

\begin{figure}
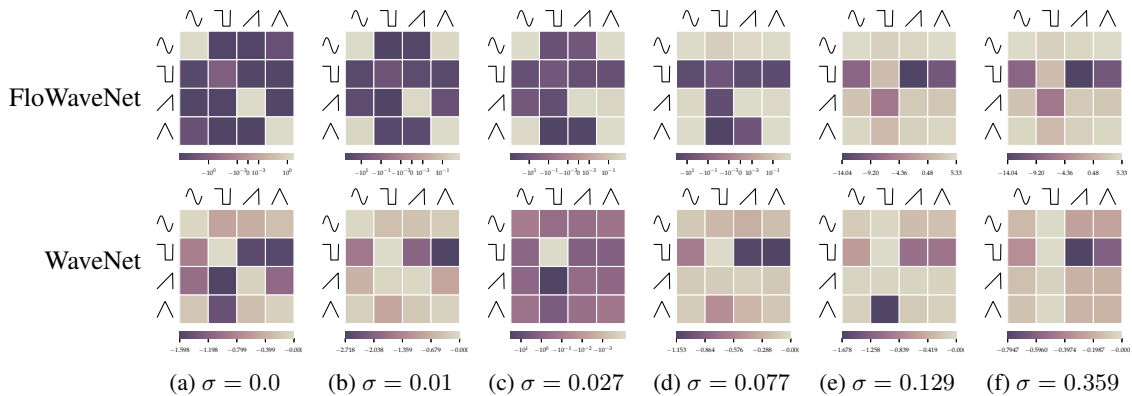

    \centering
    \adjustbox{minipage=0em,valign=t}{}%
\put(20,38){\makebox[0pt][r]{FloWaveNet}}%
\put(20,-23.5){\makebox[0pt][r]{WaveNet}}%
\hspace{1.8em}
\foreach\level in {0, 01,027,077,129,359}{
    \begin{subfigure}{0.145\textwidth}
        \includegraphics[width=\textwidth]{toy_noise_\level/channels_hm}\\%
        \includegraphics[width=\textwidth]{toy_noise_\level/wn_channels_hm}%
        \caption{\(\sigma = 0.\level\)}%
        \label{fig:noised_channels_\level}%
    \end{subfigure}
}%

    \caption{
        The cross-likelihood of the toy source channels under each model after conditioning the distribution on different levels of noise. The data noise level is going down and the conditional is going to the left. At the top is the FloWaveNet model and at the bottom of the WaveNet model.
    }%
    \label{fig:noised_channels}%
\end{figure}

Instead of retraining the deep generative priors we fine-tune~\cite{yosinski_how_2014} the noise-free models used in Section \ref{sec:discriminative}. We follow~\cite{jayaram_source_2020} in evaluating the noise-conditional model at different levels of noise $\sigma \in \{0.01, 0.027, 0.077, 0.129, 0.359\}$. In Figure~\ref{fig:noised_channels} we show the cross mean log-likelihood for increasing noise-conditionals for the Toy data dataset. We find that even with small levels of conditional noise added the discriminative power of the learned generative models decays significantly. While being smooth the noise-conditional distributions cannot be used for source separation as intended.

\subsection{Random and constant inputs}

\begin{table}
    \centering
    \begin{tabular}{llrrrr}
\toprule%
              &            &      sine &   square &      saw &  triangle \\
input         & \(\sigma\) &          &          &          &           \\\midrule%
0.0           & 0.0        &  4.8e+00 & -7.0e+02 &  4.4e+00 &   1.8e+00 \\
              & 0.359      & -5.0e-01 & -3.1e+00 &  5.1e+00 &  -2.0e+11 \\
\N(0, 0.5)    & 0.0        & -2.7e+13 & -3.4e+09 & -1.4e+05 & -1.1e+11 \\
              & 0.359      &  4.4e+00 & -5.8e+01 &  5.1e+00 & -3.8e+05 \\
\bottomrule%
\end{tabular}

    \caption{FloWaveNet: The mean log likelihood of a full receptive field of constant inputs \(\{0,1\}\) for the noise-free and the widest noise-conditioned model.}%
    \label{tab:toy_const}%
\end{table}

Previous works~\cite{sonderby_amortised_2017}\cite{van_den_oord_parallel_2017}\cite{nalisnick_deep_2019} have pointed out that generative models tend to assign high likelihood values to constant inputs. We find the same holds true for generative priors trained on the Toy data dataset. Table~\ref{tab:toy_const} shows that for the noise-free model a constant zero input is highly likely, except under the square wave prior, which we assume stems from the square wave never having the value 0.0. When fine-tuning the model with a noise-conditioning of \(\sigma=0.359\) the constant zero input becomes less likely for the sources \textit{sine} and \textit{triangle} but more likely for \textit{saw} and \textit{square}. To test whether the noise-conditioning results in simple constant inputs being unlikely but pure noise input in return becoming likely we evaluate the likelihood of noise drawn from a zero-centered Gaussian with reasonable wide variance for both the noise-free and noise-conditional model.

In Table~\ref{tab:toy_const}, we see that for the noise-free prior model, a high variance input noise sampled from \(\N(0, 0.5)\) is highly unlikely. Evaluating the same input noise under the wider noise-conditioned prior model the input becomes more likely. For the sine and saw waveform the noise input is even as likely as a normal in-class input.

We read these results to support the previous interpretation that even a small amount of noise fine-tuning can have severe effects on the estimated density. The noise-free prior models have sharp likelihood peaks around true data, in which even small amounts of added noise are highly unlikely. The noise-conditioning of the flow models flattens these peaks in so far that noise and out-of-distribution samples become highly likely, even at small levels of noise-conditioning.

\section{Conclusion}%
\label{sec:conclusion}
In this work, we show that contemporary generative models for modeling of audio signals also exhibit strong problems with out-of-distribution data as similarly described in \cite{nalisnick_deep_2019} for models of image data. Our experiments reinforce a suspicion that was also experimentally found in prior work on image data. Current deep generative models do not learn a density that is discriminative against out-of-distribution samples. We show that in our case the models lose their ability to detect out-of-distribution samples when trained with additive noise which is added to smooth the learned densities. Therefore our work further demonstrates to be cautious when applying current flow-based models to data outside close bounds of their training distribution.

\bibliography{references}
\bibliographystyle{iclr2020}
\medskip

\newpage
\section{Appendix}
\label{sec:appendix}
\subsection{Source separation with SGLD}
For better understanding of the source separation approach we had in mind using the generative models as prior we give the implementation in Algorithm~\ref{alg:langevin_sampling}.

\begin{algorithm}
    \begin{algorithmic}[1]
        \For{\(t=1\dots T\)}
            \For{\(k=1\dots N\)}
                \State\(\epsilon_t \sim\N(0, \1)\)
                \State\(\Delta\s_k^t \gets\s^t + \eta \cdot \nabla\log{p(\s^t)} + 2\sqrt{\eta}\epsilon_t\)
            \EndFor%
            \For{\(k=1\dots N\)}
                \State\(\s_k^{t+1} \gets\Delta\s_k^t -\frac{\eta}{\sigma^2}\cdot [\m  - \frac{1}{N}\sum^N_i \s_i^t]\)
            \EndFor%
        \EndFor%
    \end{algorithmic}
    \caption{The Langevin sampling procedure for source separation is fairly straight forward. For a fixed number of steps \(T\) we sample we take a step into the direction of the gradient under the priors and the gradient of the mixing constraint while adding Gaussian noise \(\epsilon_t\).}%
    \label{alg:langevin_sampling}%
\end{algorithm}

\subsection{Model and training details}
\begin{figure}
    \centering
    \begin{tikzpicture}[every node/.style={scale=0.8}]
        \tikzstyle{box}=[rectangle,draw,minimum width=25mm,minimum height=5.55mm,fill=yellow!10]
        \tikzstyle{hbox}=[box,minimum width=12mm,fill=yellow!10]
        \tikzstyle{label}=[rotate=90,yshift=6]

        \begin{scope}
            \node (z) {\z};
            \coordinate[below=10pt of z] (belowz) {};
            \coordinate[below=10pt of belowz] (aboveflow) {};
            \node[below=7pt of aboveflow,box,fill=orange!40] (flow)    {Flow};
            \coordinate[below=7pt of flow] (belowflow) {};
            \node[below=13pt of belowflow,box] (squeeze) {Squeeze};
            \coordinate[below=15pt of squeeze] (abovex) {};
            \node[below=5pt of abovex] (x) {\(\s\)};

            \draw[->] (x) to (abovex) to (squeeze);
            \draw[->] (squeeze) to (flow);
            \draw[->] (flow) to (belowz) to (z);

            \draw (belowz) -| (-1.7, -2) node[label] {\(\times n_b\)} [->]|- (abovex);
            \draw (aboveflow) -| (-1.07,-1.35) node[label] {\(\times n_f\)} [->]|- (belowflow);

            \begin{pgfonlayer}{background}
                \filldraw [inner sep=50pt,line width=4mm,black!10]
                (-1.25,-0.9) rectangle (1.25,-3);
            \end{pgfonlayer}
        \end{scope}

        \begin{scope}[xshift=3.5cm]
            \node (end) {};
            \node[box,below=23pt of end] (corder) {flip \I{even} and \I{odd}};
            \node[box,below=10pt of corder,fill=aqua!40] (ac) {Affine coupling};
            \node[box,below=10pt of ac] (an) {ActNorm};
            \node[below=12pt of an] (start) {\z};

            \draw[->] (start) to (an);
            \draw[->] (an) to (ac);
            \draw[->] (ac) to (corder);
            \draw[->] (corder) to (end);

            \draw[densely dotted] (-2.5, -1.125) .. controls (-1.6, -0.71) .. (-1.45, -0.71);
            \draw[densely dotted] (-2.5, -1.57) .. controls (-1.6, -3.19) .. (-1.45, -3.19);

            \begin{pgfonlayer}{background}
                \filldraw [inner sep=50pt,line width=4mm,orange!20]
                (-1.25,-0.9) rectangle (1.25,-3);
            \end{pgfonlayer}
        \end{scope}

        \begin{scope}[xshift=7cm]
            \coordinate[yshift=-10pt] (end) {};
            \node[hbox,left=1pt of end] (oodd) {\({out}_{odd}\)};
            \node[hbox,right=1pt of end] (oeven) {\({out}_{even}\)};

            \node[hbox, below=55pt of oeven] (wn) {WN};
            \node[hbox, below=15pt of oodd] (trans) {\(\frac{{\color{red}\x} - {\color{yellow}\B{t}}}{{\color{yellow}\B{s}}}\)};

            \node[hbox,below=80pt of oodd] (iodd) {\({in}_{odd}\)};
            \node[hbox,below=80pt of oeven] (ieven) {\({in}_{even}\)};

            \draw[->] (ieven) to (wn);
            \draw[->] (trans) to (oodd);
            \draw[->] (iodd) to node[label] {\(\color{red}\x\)} (trans);
            \draw[->] (wn) |- node[label,xshift=-20pt] {\(\color{yellow}\log \B{s}, \B{t}\)} (trans);
            \draw (ieven.east) -| (1.6, -1.3) [->]|- (oeven);

            \draw[densely dotted] (-2.44, -1.75) .. controls (-1.6, -0.71) .. (-1.45, -0.71);
            \draw[densely dotted] (-2.44, -2.22) .. controls (-1.6, -3.19) .. (-1.45, -3.19);

            \begin{pgfonlayer}{background}
                \filldraw [inner sep=50pt,line width=4mm,aqua!20]
                (-1.25,-0.9) rectangle (1.25,-3);
            \end{pgfonlayer}
        \end{scope}
    \end{tikzpicture}
    \caption{The building blocks for the FloWaveNet model. The model consists of \(n_b\) blocks (left). Each block consists of \(n_f\) flows (middle). In each flow we apply activation normalization, followed by the affine coupling (right), after which the binary mask for the even/odd mapping is inverted. The affine coupling layer uses a WaveNet with the \I{even} set as the input to output scaling \(\log s\) and translation \(t\) with which the \I{odd} set is transformed. The squeeze operator, \textit{squeezes} the time-dimension into the channel dimension doubling the number of channels.}%
    \label{fig:flow_network}
\end{figure}
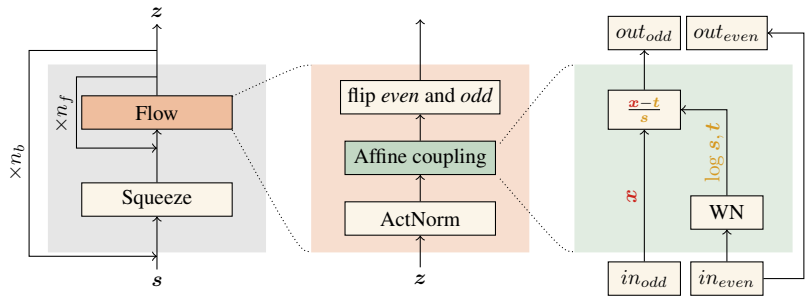

We construct the flow models closely following the architecture of FloWaveNet~\cite{kim_flowavenet_2019} which we show in Figure~\ref{fig:flow_network}. It combines the affine coupling layer proposed in RealNVP~\cite{dinh_density_2017} with the Activation Normalization proposed in Glow~\cite{kingma_glow:_2018} but does not learn the channel mixing function as in Glow and apply the fixed checkerboard masking over the channel dimension.

The WaveNets are constructed as described in the original WaveNet work~\cite{rethage_wavenet_2018}. As in the original work the outputs of the model at each time-point are modeled with a multinomial distribution with a size of 256 and therefore uses a cross-entropy loss for optimization. The quantization of the wave data is done with standard \(\mu\)-law encoding.

The hyperparameters for all for model architectures are listed in Table~\ref{tab:hyperparameters}.

The models are trained with the Adam optimizer~\cite{kingma_adam_2017}. As all models are fully convolutional the input size is in no way regimented by the architecture, only in so far that we are avoiding padding in the lower layers nevertheless we fix the size of all frames to \(2^{14} = 16384\). The initial learning rate is set to \(1e-4\) and decreased with \(\gamma=0.6\) in a fixed five-step decrease schedule. The toy model is trained with a batch size of 5 and the \texttt{musdb18} model with a batch size of 2. We train the two unconditional flows and the WaveNets are trained for each 150.000 steps. The fine-tuning with the added noise is each trained until convergence which in practice was achieved in 20.000 to 40.000 steps.

\begin{table}
    \centering
    \begin{tabular}{rccccc}
        \toprule
                                      & blocks & flows & layers & kernel size & width   \\\midrule
        WaveNet (toy)                 & 3      & -     & 10     & 3           & 256     \\
        WaveNet (\texttt{musdb18})    & 3      & -     & 10     & 3           & 256     \\\midrule
        FloWaveNet (toy)              & 4      & 6     & 10     & 3           & 32      \\
        FloWaveNet (\texttt{musdb18}) & 8      & 6     & 10     & 3           & 48      \\
        \bottomrule
    \end{tabular}%
    \caption{The hyperparameters for the FloWaveNet and WaveNet models. In case of the WaveNet blocks refers to the blocks as described in the original WaveNet architecture~\cite{rethage_wavenet_2018} while in the FloWaveNet the layers refer to the layers of the WaveNet in the coupling layers.}%
    \label{tab:hyperparameters}%
\end{table}

\end{document}